%% file: acmmanuscript.tex
\documentclass[manuscript,screen,nonacm]{acmart}

\AtBeginDocument{%
  }

\usepackage{cleveref}
\usepackage{booktabs}
\usepackage{multirow}
\usepackage{longtable}
\usepackage{array}
\usepackage{fvextra}
\usepackage{tcolorbox}
\usepackage{tabularx}
\usepackage[framemethod=tikz]{mdframed}      % prompt boxes (split across pages)
\usepackage{enumitem}      % itemize[leftmargin=*,nosep] in the worked example
\usepackage{hyperref}      % \url in the worked example
\usepackage{float} 
\usepackage{xltabular}

\usepackage{siunitx}
\newcommand{\CI}[2]{%
  \begingroup
  \setlength{\tabcolsep}{0pt}%
  [\begin{tabular}{@{}S[table-format=+2.1]@{,\;}S[table-format=+2.1]@{}}
    #1 & #2
  \end{tabular}]%
  \endgroup
}

\input{text/env}

\begin{document}

\title{Web search in Latin America: language models use less local and diverse news than search engines}

\author{Francisco W. Kerche}
\email{francisco.nunes@oii.ox.ac.uk}
\orcid{0000-0002-9621-880X}
\correspondingauthor
\affiliation{%
  \institution{University of Oxford}
  \city{Oxford}
  \state{Oxfordshire}
  \country{United Kingdom}
}

\author{Franziska Sofia Hafner}
\orcid{0000-0003-1070-8267}
\affiliation{%
  \institution{University of Oxford}
  \city{Oxford}
  \state{Oxfordshire}
  \country{United Kingdom}
}
% \email{}

\author{Felix M. Simon}
\orcid{0000-0002-0371-4653}
\affiliation{%
  \institution{University of Oxford}
  \city{Oxford}
  \state{Oxfordshire}
  \country{United Kingdom}
}
% \email{}

\author{Luc Rocher}
\correspondingauthor
\orcid{0000-0002-9956-1187}
\affiliation{%
  \institution{University of Oxford}
  \city{Oxford}
  \state{Oxfordshire}
  \country{United Kingdom}
}
\email{luc.rocher@oii.ox.ac.uk}

\renewcommand{\shortauthors}{Kerche et al.}

\settopmatter{printacmref=false}
\setcopyright{none}
\renewcommand\footnotetextcopyrightpermission[1]{}

\begin{abstract}

AI platforms are becoming a major gateway to news and current events. This shift risks amplifying existing geographical biases that favour narratives from wealthier countries over those from the Global South. This is particularly relevant in Latin America, which accounts for the largest share of AI usage in the Global South. We examine how information about Latin America changes when users shift from conventional search engines to LLM-based search. We construct a dataset of 1,901 current events across nine Latin American countries, craft realistic queries, and submit them to three search engines and four language models. We vary query language and user location to represent local, diaspora, tourist, and foreign user scenarios. Across 49,207 requests and 8,834 unique domains, language models return a less diverse set of news sources than search engines. Across user types, the share of domestic news returned by language models is inconsistent. For local users, while search engines already return only 41.3\% local news, language models return a further 3.6 to 11.6 percentage points less. Taken together, our findings suggest that LLM-based search can reproduce and, in some settings, amplify geographical hierarchies in the reach and visibility of news sources. We release an open dataset to support multilingual and geographically diverse evaluation of LLM-based search, and highlight implications for more equitable information systems design.

\end{abstract}




\maketitle

\input{text/text}

\section*{Generative AI use statement}
We used Claude Sonnet 5.5 and Claude Opus 5.5 (in Claude Code) to assist with the development of code for implementing data wrangling and statistical analysis routines, and producing figures. The authors reviewed AI-assisted code before use. We used general-purpose AI assistants (Sonnet 5.5, Sol 6.1) and Grammarly to provide suggestions for targeted edits as well as help on LaTeX table formatting.

\bibliographystyle{ACM-Reference-Format}
\bibliography{references}

\appendix
\crefalias{section}{appendix}
\crefalias{subsection}{appendix}
\crefalias{subsubsection}{appendix}
\counterwithin{table}{section}   % numbers as A.1, A.2, B.1...
\counterwithin{figure}{section}  % same for figures, if you have any

\input{text/appendix}

\end{document}

%% file: text/env.tex
\usepackage{fvextra}   % loads fancyvrb, provides breaklines
\usepackage[framemethod=tikz]{mdframed}  % tikz needed for roundcorner

\mdfdefinestyle{promptstyle}{
  skipabove=\baselineskip,
  skipbelow=\baselineskip,
  backgroundcolor=gray!4,
  linecolor=gray!45,
  linewidth=0.5pt,
  roundcorner=2pt,
  innertopmargin=6pt,
  innerbottommargin=6pt,
  innerleftmargin=6pt,
  innerrightmargin=6pt,
  frametitlefont=\bfseries\small,
  frametitlebackgroundcolor=gray!15
}

\newenvironment{promptbox}[1][Prompt]
  {\VerbatimEnvironment
   \begin{mdframed}[style=promptstyle, frametitle={#1}]
   \begin{Verbatim}[breaklines=true,fontsize=\footnotesize]}
  {\end{Verbatim}\end{mdframed}}
\makeatletter
\AtBeginEnvironment{promptbox}{%
  \let\@vspace\@vspace@orig
  \let\@vspacer\@vspacer@orig}
\makeatother

%% file: text/text.tex
\section{Introduction}\label{introduction}

Latin America accounts for a significant share of global AI use, but has little autonomy and representation in how these systems and platforms are designed. The region holds 14\% of global traffic to AI platforms~\citep{nu_cepal_indice_2025}, making it the third-largest market by consumer presence. AI use among adults in Latin American countries grew by over 20\% between 2025 and 2026, a relative growth similar to high-income countries in Europe and North America \cite{mathieu_data_2023}. This positions the region as a fruitful area to observe the effects and particularities of implementing artificial intelligence systems in regions far from where they were originally built. 

As the use of LLM-based assistants grows, these services increasingly act as a gateway to local information and news, complementing traditional search engines and integrating web search features into their interfaces~\citep{weinbrand_search_2026}. Using LLM-based assistants to seek information rather than only to produce content is now one of the main use cases in many countries worldwide \citep{simon_generative_2026}. News consumption through conversational AI systems remains limited but is steadily growing, especially among younger people~\citep{egan_digital_2026}. As \citet{egan_digital_2026} find, 10\% of people across 48 countries now use AI chatbots for news, up from 7\% last year, with 16\% of under-35s reporting that they do so. 

When users turn to LLM-based assistants to learn about events in their countries, the assistants can use web search to retrieve current information and supplement their responses. However, these replies can rely heavily on international sources. Web search is particularly relevant when a query requires up-to-date or precise information, falls outside an LLM's training cutoff, or explicitly requires external documents~\citep{lewis_retrieval-augmented_2020}. In these settings, the information presented to users depends not only on the underlying LLM but also on the sources crawled, selected, and integrated into the assistant's response. Search engines themselves already underrepresent sources from countries with lower levels of economic development~\citep{ballatore_digital_2017, cobos_origin_2021}. As a result, people in lower-income countries may be more exposed to international narratives than locally produced ones, even when international coverage is less nuanced or less attentive to local, lived realities. LLM-based search may reproduce these disparities and, through selection and synthesis, even exacerbate them.

If LLM-based search (or Generative Search) relies heavily on international sources, it may further reduce the visibility of national voices and perspectives for local users in the Global South. Studies have shown how language models portray regions of the world unevenly, often reproducing colonial or stereotypical associations to the detriment of these developing or underdeveloping regions~\citep{kerche_silicon_2026}. For example, LLMs have been found to be more accurate on national-level indicators such as GDP, Gini, and unemployment rates for countries with stronger economic indicators \citep{mirza_global-liar_2024}; to associate likeability or attractiveness rates for countries with better socioeconomic conditions~\citep{manvi_large_2024,manor_what_2025}; and to use more positive language when describing richer countries~\citep{georgiou_chatgpt_2025}. LLM-based search might further amplify these disparities if domestic sources are under-represented in Global South countries.

To examine this mechanism directly, we conducted a systematic comparison of source retrieval across search engines and LLMs for current events in Latin America, varying both query language and user location. To produce a representative set of events, we used the Global Database of Events, Language and Tone (GDELT), which gathers and classifies news coverage from across the globe. To ensure that LLM responses use web search tools, we selected recent events that fell outside the knowledge cutoffs of all evaluated LLMs across nine Latin American countries. We queried three search engines (Google, Bing, Yahoo) and four LLMs (GPT, Claude, Perplexity, Grok) using about 1,901 events involving these countries. We used domains to compare the distribution of news sources across location, language, and platform, allowing us to evaluate how LLM-based search changes the visibility of domestic and international sources relative to conventional search. 

We compare two main aspects of search engines and LLM-based assistants to understand their impacts for Latin American information environments. First, we measure the diversity of the sources they return and, second, we evaluate the prevalence of domestic news sources in responses.

We find that LLM-based search reproduces and, in some settings, amplifies inequalities already present in conventional search for news and current events in Latin America. Specifically, compared with search engines, LLMs (a) generally return a less diverse set of news sources across query language and user location; and (b) expose local users searching in their own language to less domestic news. These patterns risk reinforcing global disparities in access to information in Global South countries. We argue that LLM-based search can narrow the range of sources and perspectives available to users in Latin American information environments.

\section{Related Work}

Both search engines and, increasingly, language models function as information gatekeepers: they shape what users are exposed to, which sources become visible, and which perspectives and narratives are made easier or harder to access. Online visibility is not evenly distributed, with platforms algorithmically shaping which information is available to users \citep{cardon_dans_2013}. Search engines organise information through rankings that tend to privilege major economies and countries in the Global North \citep{ballatore_digital_2017}. Since language models use APIs from these engines for real-time inference, they risk reproducing this pattern in their search results.

\subsection{Democratic relevance of news diversity}

When language models act as mediators for news and information access, they directly affect users' capacity to understand and even critically assess their local realities. While the exact role of news in democracy is contested both normatively and empirically \citep{nielsen_one_2017}, a large body of media research has found that news helps people become more knowledgeable about politics and public affairs \citep{aalberg_how_2012} and, in some cases, more resilient against misinformation \citep{humprecht_resilience_2020, altay_news_2024}. However, this partly depends on how news and information are provided, selected, and consumed.

Changes in media market structure and media consumption can lead to widening gaps between, for example, the information “rich” and the information “poor” \citep{aalberg_media_2010}. Exposure to more “hard news” (e.g. public service broadcasting, upmarket news brands) is often associated with increased factual political knowledge compared to exposure to soft news (e.g., tabloid news) \citep{fraile_not_2014}. This issue is even more pronounced on the local level. Declining domestic political news coverage---and, by extension, consumption---has been linked to lower knowledge and participation \citep{hayes_as_2015, hayes_decline_2018} as well as increases in domestic corruption \citep{matherly_no_2024}. How well people are informed about public affairs might therefore become increasingly mediated by AI systems's selection, ranking, and presentation of information.

Access to diverse information is not simply a technical property of information systems, but a democratic and epistemic need. Media pluralism makes it harder for dominant actors to monopolise narrative framing \citep{baker_media_2006}. The internet and World Wide Web were initially seen as opportunities to expand this plurality, but deregulation and market concentration have often resulted instead in the marginalisation of dissenting or local perspectives \citep{horwitz_media_2007}. When search engines present both international and domestic sources on local issues, they yield a variety of viewpoints. At times, international media can present narratives aligned with multinational institutions, even opposing mainstream local news \citep{araujo_framing_2021}. On the other hand, failing to represent local narratives can overlook diverse voices and reduce nuance in portraying these realities. Given that democratic participation partly depends on access to diverse and locally relevant information, systems that mediate this access should be evaluated not only for their accuracy, but also for whose sources, geographies and perspectives they privilege.

\subsection{Source concentration and geographical bias in search engines}

\paragraph{Source concentration} Search engines are often treated as neutral entry points to the web, but their ranking systems organise visibility in ways that privilege some sources over others. In search engines, online relevance metrics serve as proxies for authority. This can produce two main changes in social organisation. First, institutions (e.g. media outlets, political organizations, intellectual centres) have to increasingly compete for online validation in order to maintain their previously established forms of prestige and recognition \citep{cardon_a_2015}. Second, search engines tend to centre voices that do not deviate strongly from mainstream narratives, over-representing dominant groups and perspectives \citep{otterbacher_competent_2017}. Empirical work confirms this pattern: while individual search outputs may contain relatively diverse sources, aggregate results across queries tend to converge on a smaller set of high-traffic domains \citep{hernandes_auditing_2024}, effectively amplifying dominant voices rather than broadening exposure. 

At the same time, search engines should not be understood only as diversity-reducing systems. Search engines can, for example, offer greater source diversity compared to social media or direct news consumption\citep{fletcher_automated_2018, fletcher_more_2023}. This suggests that search engines occupy an in-between position: while they may increase exposure relative to some forms of media consumption, they still concentrate attention on a limited set of already visible actors and narratives.

\paragraph {Geographical bias} Sources can also be geographically concentrated. Domestic news coverage correlates strongly with countries' economic development and publishing infrastructure \citep{ballatore_digital_2017}. On the home page of Google News for different countries, Brazil, Mexico, and Colombia present, respectively, 88.9\%, 78.3\%, and 63\% domestic sources on the platform's main page, with other content being from international venues, portraying how news aggregators can reduce the visibility of domestic sources in countries with lower media presence \citep{cobos_origin_2021}. These patterns reflect a broader structural inequality in the web itself. More than 78\% of global domains are registered in the United States and Europe, a pattern that is reproduced in participatory platforms, such as Wikipedia edits and GitHub \citep{graham_geographies_2022, graham_uneven_2014}. As a result, web search can privilege international narratives, especially for countries with weaker publishing and media infrastructures and lower levels of economic power. Because many web-enabled language models rely on search engine APIs, these existing inequalities are likely to be reproduced in LLM-generated answers. 

\subsection{Language models search differently than search engines}

Search engines and language models differ in how they search, organise and present information. \textit{Search engines} are typically prompted with short keyword-based inputs, which are then converted into context-dependent keywords and matched against indexed web sources in their databases \citep{xu_chatgpt_2023}. They search for both entities and context in order to rank sources by relevance. \textit{Language models}, by contrast, can process longer conversational inputs and restructure users' queries into multiple searches \citep{sharma_generative_2024, wazzan_comparing_2024}. They do not index the web themselves, but often interact with these results through search engine APIs. The ranked lists they receive can then be presented in different orders and positions, according to internal criteria, including relevance for paywalled content or elevating visibility for trusted domains \citep{he_zero-indexing_2024}.

The output users see also differs significantly. \textit{Search engines} provide a ranked list of results, and these rankings strongly shape where a user might click \citep{ulloa_search_2024, haas_ranking_2017}. \textit{Language models}, by contrast, produce synthesised natural-language responses. \citet{worledge_extractive-abstractive_2026} discuss how LLMs cite sources indirectly, inconsistently, or not at all, reducing user interaction with sources as a whole; however, they also note that greater abstraction from original sources improves content comprehension, at the cost of transparency.

Because of how language models respond, users are less likely to click links or even see which sources informed their replies. This can make the geography of sources less visible to users even as it may continue to shape the information they receive.

\subsection{Geographical and cultural bias in language models} 

Language models may inherit geographical inequalities, but are also prone to introducing their own forms of geographical and cultural bias. LLM training corpora are heavily skewed toward the Global North and can undermine Global South perspectives, or even present stereotypical views and associations \citep{kerche_silicon_2026, baack_critical_2024, helm_diversity_2024}. Researchers have measured these geographical preferences through experiments that compare the scores, choices, or the narratives models produced for different locations. \citet{manvi_large_2024}, for instance, asked LLMs to rate locations from 0 to 10 on attributes such as attractiveness and likeability, describing each location only by its coordinates and its distance to ten nearby landmarks; in that scenario, models assigned higher ratings to locations in higher-income countries. \citet{georgiou_chatgpt_2025} observed a similar preference in longer generated texts, and found that models used a higher proportion of positive words across countries with a high Human Development Index (HDI).

Geographical inequalities are also present in hiring and employment-related decisions. Researchers found that language models exhibit nationality bias, disadvantaging participants from certain countries in selection processes. \citet{salinas_unequal_2023} varied nationality and gender when asking for GPT-3.5-turbo and LLaMA-65B to suggest jobs and salaries for a laid-off job seeker in the United States. Recommendations varied by nationality, especially for Mexican job seekers, who received recommendations for lower-paying jobs and the lowest median salaries of all 20 nationalities tested, a pattern most pronounced in ChatGPT. These findings suggest that geographical disparities have real-world consequences when incorporated into decision-making processes.

Another issue is how language models can reshape travel destinations, potentially reducing tourism and harming local economies. When asked for travel recommendations, models tend to over-represent already highly sought-after locations, reinforcing structural imbalances in touristic destinations \citep{andreev_destination_2025}; when providing travel recommendations for countries with lower GDP, language models also tend to be less unique, with fewer locations and less detail. The preference for highly sought-after locations and high-income locations is also reproduced at the city level for travel destinations, starting a business, or even moving \citep{dudy_unequal_2025}.

One way to reduce knowledge disparity in models for Global South contexts is to train multilingual models on data that better reflects local realities worldwide. However, research suggests that this is typically insufficient to improve local knowledge or even provide more favourable descriptions of these countries. First, asking multilingual language models in local languages does not necessarily ensure they match culturally expected norms; for instance, when asked about the colour of a wedding dress in Chinese (where the traditional wedding dress colour is red), they can frequently reply white, reflecting Anglo-centric norms of expression \citep{yin_geomlama_2022}. They have also been shown to cite entities, places, events and cultural products from the United States more often than from any other country, even when queried in a non-English language \citep{wang_not_2024}. Conversely, when provided with information about a country and asked to explicitly name it, the model is less likely to accurately identify countries from the Global South or the Global East than those from the Global North and Global West \citep{lalai_world_2025}. Finally, measures that assess the extent of this issue can sometimes underestimate it, since researchers often use translated benchmarks to validate multilingual capabilities, thereby overstating the relevance of Western knowledge and standards while overlooking locally relevant aspects \citep{etxaniz_bertaqa_2024}.

\paragraph{Web search in language models} 

Another way to reduce geographical knowledge disparities in LLMs is to let them rely on external knowledge. In regions where models have less knowledge, web retrieval might improve their local knowledge. \citet{lertvittayakumjorn_towards_2025} compare three groups of models: models without web access; Gemini with Google web search; and models with access to a purpose-built cultural knowledge base. Search grounding significantly improved Gemini's accuracy – although with a small effect size, and the cultural knowledge base presented mixed effects. However, while slightly more accurate, this made models more likely to select stereotypical answers on stereotype-avoidance tests, and it did not increase cultural familiarity for open-ended stories.

Although language models might be able to access information in local languages to improve their understanding of these realities, this is still not sufficient. When dealing with multiple languages in their search results, language models reduce the quality of their output. \citet{liu_xrag_2025} used documents in multiple languages covering recent news in English, German, Spanish, Chinese, and Arabic, ensuring that language models would have to rely on these documents to reply. Without translating documents into English, models' ability to answer specific questions decreased by 4 percentage points (57.6\%-61.6\%), showing limitations in working across multiple languages. Even when relevant information is available, models may fail to retrieve, integrate, or represent it in ways that reflect local knowledge and perspectives. The problem is, therefore, not only whether language models can access specific knowledge, but also whether they select and synthesise sources in ways that preserve geographical diversity.

\paragraph{Source diversity in language models} As language models can search the web to increase their knowledge, diversity of information can have a significant impact on providing a more comprehensive understanding of issues and events. Research on language model summarisation indicates that they tend to reduce rather than increase informational diversity. Language models' summaries tend to align more closely with users' prior beliefs, reinforcing those views, whereas summaries expressing opposing views tend not to alter users' original perceptions \citep{sharma_generative_2024}. Language models can also flatten the plurality of perspectives present in source documents to a single viewpoint \citep{narayanan_venkit_search_2025}. This is particularly important in news and public affairs, where disagreement and uncertainty, as well as competing interpretations, are often central to democratic participation. Overall, less is known about how web-enabled language models select, cite, and synthesise sources when responding to queries about countries outside dominant information centres, such as those in Latin America. Therefore, understanding how LLM-mediated search reproduces, mitigates, or amplifies the geographical inequalities already present in search engines is fundamental for measuring the impact of these systems on how search will be used in the Global South.

\section{Methods}\label{methods}

\begin{figure}
    \centering
    \includegraphics[width=1\linewidth]{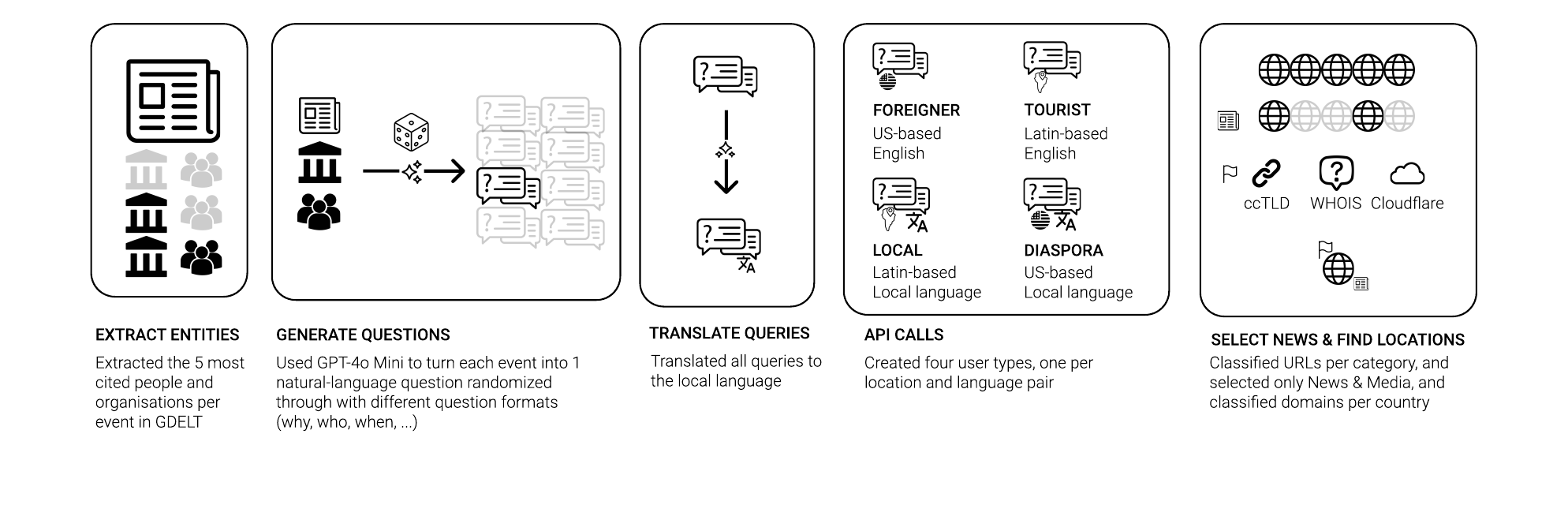}
    \caption{\textbf{Methodology for data extraction and classification of sources.} Steps used to construct the dataset from GDELT events, generate and translate queries, retrieve results with user types, and classify news sources by location.}
    \Description{Five boxes showcasing the steps for the methodology, accounting from data extractino, query generation, query translation, API calls and news selection.}
    \label{fig:methods_card}
\end{figure}

We conducted a comparative study of the sources returned by three conventional search engines and four language models for recent events in Latin America. We followed four steps: event selection, query construction, content categorisation, and source-origin classification. Overall, we included 1,901 events in nine countries and, by varying query language and user location, produced 49,207 requests in total. \Cref{fig:methods_card} shows the full methodology.

\paragraph{Event Selection} We selected events from the Global Database of Events, Language, and Tone (GDELT), which collects and groups news coverage from around the world into events based on the entities, places, and event types mentioned. We used events from the 29th of April to the 29th of May 2025, after the knowledge cutoff of all selected models, increasing the likelihood that responses require searching the web.

We selected the nine Latin American countries most frequently mentioned in GDELT (Brazil, Mexico, Argentina, Colombia, Chile, Peru, Venezuela, El Salvador, and Haiti). For each country, we extracted the 200 most-mentioned and 200 least-mentioned events, and excluded events with fewer than five mentions to ensure events were sufficiently relevant. Overall, this produced an initial corpus of 3,600 events.

The first author then reviewed each event headline and removed events that could not be properly associated with the country in question, such as events centred on institutions or individuals from outside the country. The final corpus contains 1,901 events: 1,130 involving only one country and 771 involving more than one country.

\subsection{Query construction}\label{methods:query_construction}
For each event, we used GPT-4o-mini-2024-07-18 to generate one question (see \Cref{subsec:headline-question-gen} for the prompt). To reduce sensitivity to question format, we instructed the model to events to eight distinct question openings (What, Who, Why, How, When, Where, Is, Could), then randomly picked one. We then translated each question into the country's local language (Portuguese for Brazil, French for Haiti, Spanish elsewhere) (see \Cref{subsec:query-translation} for the prompt). The first author, proficient in all four languages, manually reviewed a sample of 200 questions. Of these, 93\% clearly referred to the relevant event, person, or organisation and could be understood without access to the original headline. The same author separately reviewed 100 questions, finding 97\% to be accurately translated.

We submitted queries to search engines and language models under four user scenarios, spanning two languages (English and the local language) and two locations (the United States and the relevant Latin American country). \Cref{tab:user_types} summarises the four resulting scenarios: a foreigner (US-based, English), a diaspora user (US-based, local language), a tourist (in-country, English), and a local (in-country, local language). 

We queried four language models through their APIs with web search enabled: GPT-4o-mini-2024-07-18, claude-3.5-haiku-20241022, grok-3-latest, and perplexity-sonar. We queried three search engines through the DataForSEO SERP API (to reduce costs and access search engines which do not provide APIs) (see \Cref{tab:appx_systems_settings} for model specificity). Since users are more likely to interact with the first results presented to them by search engines, and rarely beyond the first page~\citep{pan_google_2007}, we retained only the first 10 results for each search engine query  (see \Cref{tab:appx_query_examples} for a sample of the data used to generate the queries).

When supported, we set user location in the API of each relevant language model. Claude-3.5-haiku-20241022 was available only for the US, Brazil, Argentina, Mexico, Chile and El Salvador. Bing results were unavailable for Brazil via the SERP API. When comparing two models, we exclude any unavailable values and use only the remaining regions.

\begin{table}[htbp]
\centering
\caption{Definitions of user types by user location and query language.}
\label{tab:user_types}

\begin{tabular}{@{}llll@{}}
\toprule
\textbf{User type} & \textbf{User location} & \textbf{Query language} & \textbf{Interpretation} \\
\midrule
\textbf{Local}
& Event country
& Local language
& User searching about their country in the local language \\

\textbf{Tourist}
& Event country
& English
& English-speaking user searching from the event country \\

\textbf{Diaspora}
& United States
& National language
& US-based user searching in the event country's local language \\

\textbf{Foreigner}
& United States
& English
& US-based user searching in English \\
\bottomrule
\end{tabular}
\end{table}

\subsection{Classifying sources}

We treated all retrieved URLs as sources, whether or not those were directly cited in LLM responses. For GPT, for which the underlying retrieved URLs were unavailable, we used the set of sources cited in its response.
 
\paragraph{Content category}
For every retrieved URL, we classified the root domain using Cloudflare's content taxonomy (e.g. \textit{News \& Media}, \textit{Government \& Politics}, or \textit{Education}). Because of API limits and cost, classifying all URLs using Cloudflare was impractical (over U\$5,000 for all 5,767 domains). Instead, we prompted GPT-4o-mini-2024-07-18 to assign each domain to one of Cloudflare's categories using the first 3,000 characters of its landing page (see \Cref{subsec:domain-category} for the prompt). We validated this procedure against our adapted Cloudflare's taxonomy for 250 domains, achieving 89\% agreement with the original classification. The analyses below are restricted to domains classified as \textit{News \& Media}.

\paragraph{Country of origin}
We assigned each domain to a country of origin in four steps. First, we inferred the location from the country code top-level domain (ccTLD), such as .br for Brazil or .co for Colombia. Second, for domains with generic top-level domains (e.g. .com, .org, .net), we used the country of registration in WHOIS records. Third, if neither identified the domain's location, we used the country with the largest traffic share in Cloudflare Radar as a proxy for the domain’s geographical origin. Finally, we manually classified the top 500 \textit{News \& Media} domains, which account for 84.3\% of all news sources cited in our initial corpus. Overall, only 0.001\% of results could not be assigned a country using this process and were subsequently removed (see \Cref{tab:appx_query_language_by_country} for the number of events and sources).
We classified a news outlet as \textit{domestic} when its assigned country matched the country associated with the queried event.

\subsection{Analysis}

To compare web search with LLM-based search, we calculate two metrics: source diversity $N_{su}$ and share of domestic sources $D_{su}$, for each system $s$ and user type $u$. For both metrics, we estimate 95\% BCa confidence intervals from 2,000 bootstrap resamples.

\paragraph{Source diversity.}
We aim to capture how a system distributes attention across news outlets, and not just how many outlets appear in a response. A system may return many outlets each time but still rely exclusively on the same outlets across events. We call this \textit{source concentration}: concentration is higher when different responses draw from the same outlets, and lower when they draw from a wider range of outlets. For each system and user type, let $S_i$ denote the news outlets returned in response $i$. We measure concentration using the collision probability,
\[C_{su} = \frac{1}{n(n-1)} \sum_{i\neq j} \frac{|S_i\cap S_j|} {|S_i|\,|S_j|},\]
with $n$ the number of responses with at least one news outlet. We report its inverse, the \textit{effective} number of news outlets $N_{su}=C_{su}^{-1}$. Higher values indicate that attention is distributed across outlets, and lower values indicate reliance on a smaller set.

\paragraph{Share of domestic sources.}

For each response, we calculate the proportion of retrieved outlets classified as domestic. For each system $s$ and user type $u$,
\[D_{su} = \frac{1}{n} \sum_{i=1}^{n}\frac{|S_i^{\mathrm{dom}}|}{|S_i|},\]
with $S_i^{\mathrm{dom}}\subseteq S_i$ the outlets whose country of origin matches the country associated with the event.

\section{Results}\label{findings}

We empirically compare how four language models and three search engines retrieve news sources for recent events in Latin America. For each system, we test four conditions that vary query language and user location:  searching in the local language from the event country (local), in the local language from the United States (diaspora), in English from the event country (tourist), and in English from the United States (foreigner). Across these conditions, we evaluate the diversity of retrieved news sources and the share of sources from the country where the event took place.

\subsection{Language models reduce information diversity}

\begin{figure*}
    \centering
    \includegraphics[width=1\linewidth]{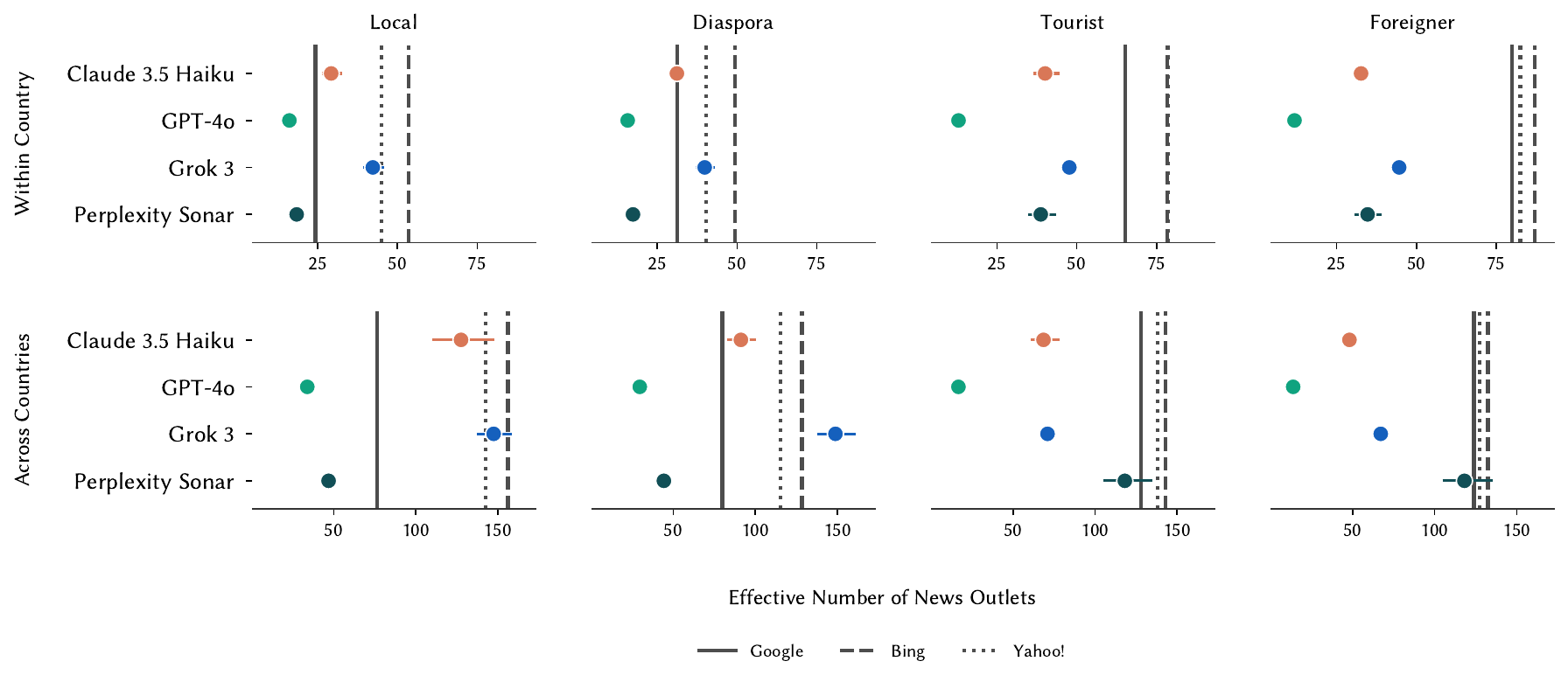}
    \caption{\textbf{Effective number of news outlets from language models compared to search engines} within countries (top) and across countries (bottom), by user type. Higher values indicate greater source diversity, while smaller numbers indicate greater concentration. Error bars show 95\% confidence intervals.}
    \Description{A two faceted dot plot with 95\% confidence intervalls. The columns are
four user personas: Local, Diaspora, Tourist, and Foreigner. The x-axis is the
effective number of news outlets, while the y-axis has the language model. Three lines represent search engines: Google (solid), Bing (dashed) and Yahoo! (dotted).}
    \label{fig:entropy}
\end{figure*}

\Cref{fig:entropy} shows the effective number of news outlets $N_{su}$ returned by each system. It measures how concentrated a system's sources are across queries (see \Cref{tab:appx_outlet_collision_by_user} for the set of full results). For instance, if $N_{su}=100$, this does not mean the system returned 100 sources, but that sources from the system overlap as much as if they were drawn evenly from a sample of $100$ outlets. Higher values therefore indicate that queries draw from a broader set of news outlets, and lower values indicate that the same outlets repeat across queries more often, meaning they are drawn from a smaller set. We calculate $N_{su}$ at two levels: (a) within countries, to explore how concentrated news sources are when these systems report on local events, and (b) across countries, to explore how concentrated in global dominant media the overall reporting in Latin American contexts is.

Importantly, search engines can retrieve more diverse news sources for some user conditions compared to others. In within-country comparisons, local users receive 42 fewer effective news outlets on average than foreigners. This effect is more pronounced for Google, with 55.6 fewer effective outlets. Across countries, the same pattern holds only for Google, which returns 76.5 effective outlets for locals and 124 for foreigners; while Bing and Yahoo! vary much less across conditions. This suggests that users who do not search the web in English already experience a reduced diversity of news outlets from search engines.

For queries about events in the same country, \Cref{fig:entropy} shows mixed results for local-language search, while English-language search is uniform. For searches in local languages (local and diaspora user conditions), Grok provides consistently higher effective news outlets, while Claude has higher diversity in locals and no effect for diaspora users. GPT-4o and Perplexity Sonar return fewer effective news outlets than search engines, regardless of user conditions. The reduction is more pronounced for queries in English: for both tourists and foreigners, every language model returns fewer effective news outlets than search engines. Source diversity remains reasonably stable for each language model across conditions, suggesting that much of this gap comes from the greater diversity search engines already provide for tourists and foreigners in English.

For queries across countries, we observed mixed effects between language models and search engines (\Cref{fig:entropy}). For local-language queries, GPT-4o and Perplexity return fewer effective news outlets than search engines; Grok returns more than search engines, and Claude shows mixed results. For English queries, we observe a more consistent pattern: all language models except Perplexity show a significant reduction in information diversity. Unlike the other models, Perplexity does not appear to personalise responses based on user location. Overall, with the exception of Perplexity, English-speaking users of LLM-based search will receive news from a less diverse ecosystem, across all tested language models in any user condition.

\subsection{Language models use fewer domestic news sources for local users}

\begin{figure*}
    \centering
    \includegraphics[width=1\linewidth]{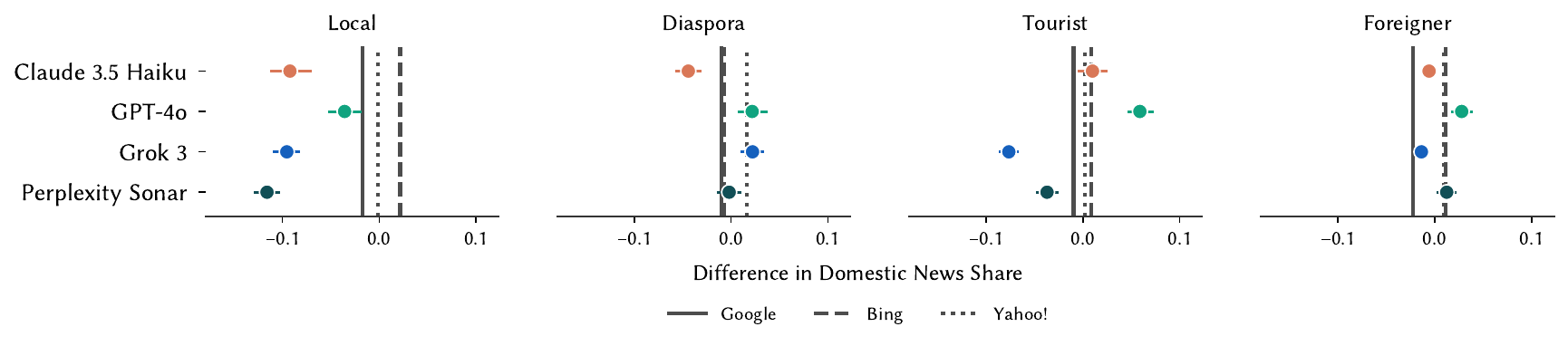}
    \caption{\textbf{Difference in share of domestic news between language models and search engines,} by user type. We report the difference, in percentage points, in $D_{su}$ for each language model relative to the average search engine. Error bars show 95\% confidence intervals.}
    \Description{Four plots with dot plots where each point represents a language model. Each model has a colour and a bar for the confidence interval. Three vertical lines represent search engines: Google (solid), Bing (dashed) and Yahoo! (dotted).}
    \label{fig:simple_news}
\end{figure*}

\Cref{fig:simple_news} shows the share of domestic news outlets returned by each language model, relative to the average search engine. Negative values indicate that a language model returns a smaller share of domestic outlets, positive values a larger share. We also display separate lines for each search engine, with Google returning the smallest share of domestic outlets for all user types.

For local users, all language models return a smaller share of domestic news than the average search engine. The reductions range from 3.58 percentage points for GPT-4o to 11.6 percentage points for Perplexity. Similar results appear if comparisons are made only to Google, the most widely used search engine, with only GPT-4o not showing a significant reduction. For the other three user conditions, the effect depends on the model as well as the user location and query language. Overall, local users face the most significant effects on domestic news when switching to LLM-based search.

\begin{figure*}
    \centering
    \includegraphics[width=0.9\linewidth]{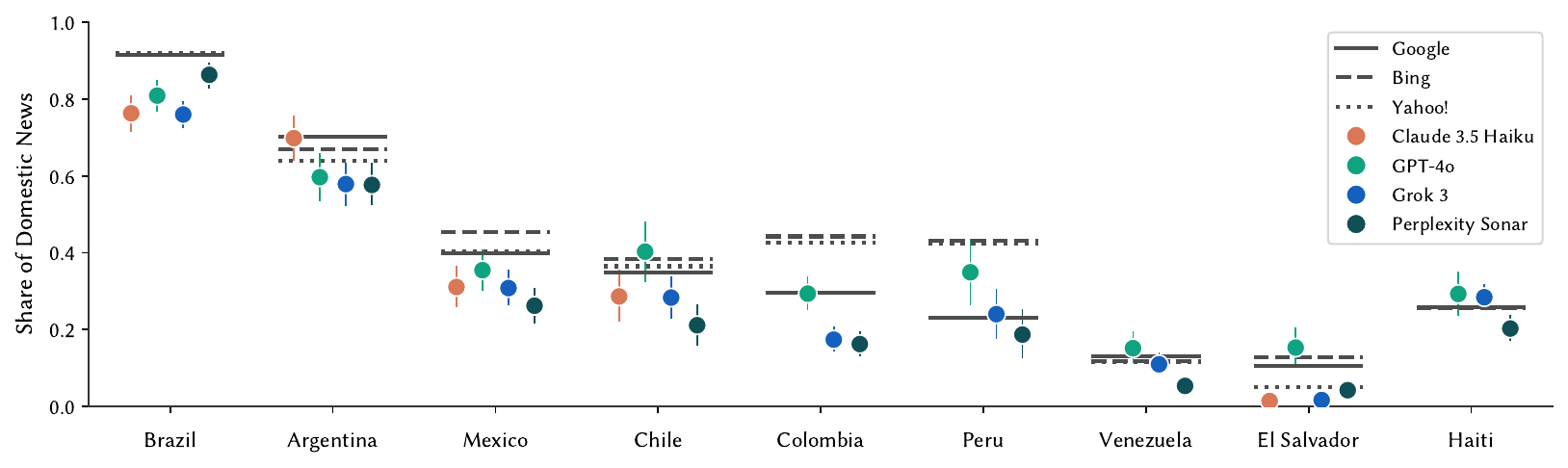}
    \caption{\textbf{Share of domestic news per country for local users.} Absolute share of domestic news for search engines and language models, broken down by country and ordered by the number of registered ccTLD domains. Error bars show 95\% confidence intervals.}
    \Description{Dotted plot with countries on the x-axis and four points per country. Points represent the share of domestic news per language model. Three horizontal lines represent search engines: Google (solid), Bing (dashed) and Yahoo! (dotted).}
    \label{fig:full_image}
\end{figure*}

We then examine if the reduction of domestic news observed for local users is consistent across countries. \Cref{fig:full_image} shows the share of domestic news $D_{su}$ for local users per country and system (see \Cref{tab:appx_simple_news_vs_average_models} for the set of full results). We find that $D_{su}$ varies substantially across the nine countries: only Brazil and Argentina have an average domestic news share above 50\% for all language models and search engines. Across language models and search engines, we find that $D_{su}$ generally follows the number of registered ccTLD domains, which we used to order countries in \Cref{fig:full_image}. This suggests that countries with more registered websites generally have a higher share of domestic news, and that the visibility of domestic sources is tied to a country's online presence.

We observe differences across search engines: while for some country pairs, such as Chile and Venezuela, $D_{su}$ is similar across search engines, the difference between Google and other search engines exceeds 20 percentage points in Peru. Language models therefore operate on top of search environments that can already vary substantially in the share of domestic news they present to users.

In all countries, language models rarely increase the share of domestic news for local users. They generally return either a smaller share or no change relative to search engines. GPT-4o is the main exception, with more domestic news in Peru and El Salvador compared to Google. Overall, language models tend to preserve or reduce the share of domestic news already found in conventional search engines, actively exacerbating the dominance of international sources in search engines.

\section{Discussion}\label{discussion}

Taken together, our findings suggest that LLM-based search can reproduce and further amplify information inequalities in Latin America. Compared to conventional search engines, language models generally draw from a narrower set of news outlets, especially for queries in English. They also return a smaller share of domestic news sources for local users, searching from within their country in their local language.

\paragraph{Location and language shape source inequality.}

Claims about LLMs often assume that these systems behave identically regardless of where they are being used. For web search, however, both location and language shape which sources are available, retrieved, and selected. This is particularly important for news and current events, where geography and language are a crucial part of the information being sought. Users expect that replies from models that discuss local events are informed by sources that are locally relevant and can provide meaningful insight on the topic ~\citep{egan_digital_2026,simon_generative_2026,chatterji_how_2025}.

These differences mirror existing concerns about linguistic variations in LLMs. Research has shown that LLM performance can vary substantially across languages, especially between high- and low-resource languages~\citep{beyene_msteb_2025,luo_gloteval_2025,ahuja_megaverse_2024}. Multilingual retrieval can similarly degrade LLM performance~\citep {liu_xrag_2025}. Our results suggest that web search, by contrast, adds an additional dimension: sources returned by LLM-based search not only vary across languages, but also location. Latin American users who search in their own language and from their own country are the most affected for the visibility of local information.

Source inequality also emerges on top of existing geographic inequalities in conventional web search. Search engines tend to return fewer domestic sources in countries with weaker economies and smaller publishing infrastructure~\citep{ballatore_digital_2017, cobos_origin_2021, cordeiro_analysis_2024}. We find that LLM-based search does not correct these inequalities. Instead, for local users who interact with language models in Latin America, all four tested language models return a smaller share of domestic news than search engines. Improvements in model generation or multilingual capability may thus not, by default, solve these issues.

As LLM-based search can amplify these inequalities, developers should conduct regular long-term monitoring to assess whether source concentration and under-representation of domestic news persist. Ideally, search results should return neither only local nor only international sources, as source diversity and international perspectives can positively affect domestic issues \citep{araujo_framing_2021}. Instead, source selection should support a balanced view of sources that can represent local narratives and produce a pluralist view of local political events. For this, independent evaluations and transparency over source usage can go a long way. Where they decline or fail to do so, this task will fall to independent researchers and regulators, who, in other areas---such as general model capabilities---already engage in regular monitoring of such systems. This is particularly relevant because language models have made source inequality dynamics harder to evaluate, since they often do not cite the sources they use, nor is it possible to indicate which information comes from which source \citep{worledge_extractive-abstractive_2026}.

\paragraph{Source concentration and visibility of domestic news.}

Source selection can affect which institutions receive visibility through an LLM-based search. Researchers have long highlighted the role of search engines as information gatekeepers \citep{introna_shaping_2000,cardon_dans_2013,nechushtai_what_2019}, and, as a result, how they affect the visibility and reputation of institutions and public figures~\citep{cardon_a_2015}. Our findings highlight how LLM-based search adds another dimension and, when it repeatedly relies on a smaller set of outlets, can further concentrate attention on already-prominent institutions.

This concentration of news and information could potentially have long lasting effects on news ecosystems. If LLM-based search directs less attention to domestic news outlets, it could send fewer users to their websites and, in turn, reduce subscription and advertising revenue \citep{zhao_impact_2026}. Addressing these risks requires understanding where, in the search pipeline, this concentration happens. Without greater scrutiny, they could undermine local information and news, which is shown to increase participation in politics~\citep{hayes_as_2015, hayes_decline_2018} and hold those in power to account \citep{matherly_no_2024}.

\paragraph{Auditing source selection across systems.}

AI companies provide limited information about how their systems conduct web search, which sources they retrieve, and how they select sources. LLM-based search is better understood as a combination of several technical systems rather than the output of a single language model. Following the notion of a technical ensemble \citep{hui_existence_2016}, the observed inequalities can emerge from interactions among model tool calls, search infrastructure, ranking algorithms, and the model's own selection of retrieved documents. Evaluations of LLM-based search should therefore examine both source retrieval and selection alongside the quality of the generated model responses.

Our study provides a practical framework for measuring how geographical bias can emerge through source selection in LLM-based search. Rather than evaluating the accuracy of the final answer, we compare the sources retrieved by LLMs and search engines while varying user location and query language. This makes it possible to identify inequalities that would be obscured when evaluating only the responses for LLM-based search, or treating them as location-independent.

Our methods are easy to reproduce and extend. Applying the same queries across multiple languages and locations requires little additional cost but helps shed light on the different sources that users encounter. Similar evaluations could be conducted in other relevant regions, for instance those with small media ecosystems, less represented languages, more or less state control over media, and where LLM-based search is becoming an important pathway to news and current events.

Separating user location from query language is important for understanding the effects of geography and language as two separate but interacting factors. Research that aims at validating the ways that information is presented to users in other regions, or how they interact in single or multi-round conversations should account for the interactions of these two variables with the low cost of altering API locations and doing query translation \citep{katsis_mtrag_2025}.

\paragraph{Governance and accountability of AI platforms.}

Our findings do not show evidence that news concentration is intentional. Instead, they suggest that AI systems, if not properly instructed or designed otherwise, can further amplify existing search inequalities that have gone largely unaddressed so far. Inversely, developers could feasibly shape source selection more deliberately to generate further visibility of domestic news sources, similar to how platforms in the past have intervened in the ranking and moderation of sources and content. 

As AI systems, including AI assistants, become central to information and news, Latin American countries could find their perspectives and narratives poorly represented because sources are concentrated in the Global North. Developers and regulators can monitor and amplify voices to build a more diverse information ecosystem and, where needed, address imbalances in representation, contributing to a more equitable and fair information ecosystem.

Recent regulations could offer researchers additional ways to study these patterns systematically and at scale. Under the EU Digital Services Act, large platforms and search engines are required to assess systemic risks arising from their systems and algorithms. The August 2026 designation of ChatGPT as a Very Large Online Search Engine requires OpenAI to identify, assess, and mitigate risks arising from how sources are retrieved and selected under Article 34(1) \citep{lemoine_what_2026}. It also, as a result, offers vetted researchers the right to request data under Article 40 to understand `systemic' risks, including to media pluralism. These obligations could be fruitful for researchers aiming to understand how sources are selected and personalised to users' locations and language at scale, especially in multi-turn interactions.

\paragraph{Limitations}

Due to the diversity of media landscapes or differences in governmental control, our findings may not generalise beyond the nine Latin American countries presented. The Global Database of Events Language and Tone (GDELT) provides, however, a convenient database across many countries to cheaply reproduce this evaluation elsewhere. Researchers can extend the audit to other regions at a marginal cost, consider other ways to stratify this distribution, or even focus on other types of source concentration, such as the internal diversity of a specific media outlet. Extending our analysis would help show whether the effects we observed in Latin America persist in countries with different news ecosystems, languages, and web presence.

Prior interactions may further affect which sources are retrieved or selected. Both language models and search engines adapt to previous interactions from users. Search engines alter rankings or autocomplete suggestions based on users' search history, and language models are more often implementing `memory' mechanisms that learn user preferences across days and weeks. Our four user scenarios capture differences in location and query language, two factors that shape which sources systems cite (\cref{methods:query_construction}), but cannot account for these long-horizon, user-specific effects. We found that using single interactions with no user history or personalisation isolates the effects of location and language, and facilitates comparison to search-engine interactions as a baseline. Research with this scope can still use single-round comparisons as a useful baseline against which personalised behaviour could be measured.

\section{Conclusion}

LLM-based search does not simply inherit existing geographical inequalities from conventional search engines. Across nine Latin American countries, we find they generally draw from a less diverse set of news outlets than search engines, particularly for foreign users and tourists. For local users searching in their own language, all four tested language models rely on a smaller share of domestic news than the average search engine.

These findings suggest that LLM-based search not only reproduces but can also amplify geographical inequalities in access to information. During the models' search pipeline---generating queries for search engines, selecting what to use and present to users, and finally, how to reply---LLMs perform numerous opaque actions that lack transparency and accountability. Although benchmarks and studies have evaluated how language models summarise news and information, how they find information is equally crucial and remains largely underexplored.

%% file: text/appendix.tex
\section{System Prompts}

\input{appendix/extra/prompts}

\section{Query design, examples and settings}

\input{appendix/extra/systems_and_settings}

\input{appendix/extra/query_examples}

\input{appendix/extra/query_design}

\input{appendix/extra/worked_example}

\section{Additional results}

\input{appendix/tables/outlet_collision_by_user}

\input{appendix/tables/simple_news_vs_average_models}

%% file: appendix/extra/prompts.tex
We used GPT-4o-mini for each of three steps in the pipeline. First, we generated natural-sounding questions from a headline and its associated GDELT entities, using a temperature of 0.7 (\Cref{subsec:headline-question-gen}). Second, we translated each question into the event country's local language, Spanish, Portuguese, or French, using a temperature of 0. Third, we classified each domain into Cloudflare's content taxonomy based on its landing page text, using the first 3,000 characters, and a temperature of 0.3. The full prompts for all three steps are provided below.

\subsection{Headline question generation}
\label{subsec:headline-question-gen}

We instructed \texttt{gpt-4o-mini} to generate a natural-sounding question using both the headline presented by GDELT, and the set of entities it presented.

\begin{promptbox}[Prompt]
You are an AI assistant that helps generate natural, search-like questions based on breaking news. Your input includes a news headline, the people and organizations named in the story, and the date when the event occurred.

Your task is to write 8 casual, realistic questions that someone might search online to understand or follow up on this story. Follow these rules:

1. Each question must start with a different word from this list: "What", "Who", "Why", "How", "When", "Where", "Is", and "Could".
2. Each question should:
   - Be natural and conversational, like something someone would actually Google
   - Use the headline and, where relevant, include one or more people or organizations
   - Include the date of the event to give temporal context
   - Be concise, clear, and non-redundant
   - Assume the reader has *only* seen the headline, people/orgs, and date
3. Present one question per line in a numbered list (1–8).
4. Do **not** repeat the headline or provide explanations.

Remember to keep the language informal and conversational. Avoid using jargon or overly complex terms unless they're essential to understanding the event.

Each question should be self-contained — avoid vague references like “the new law” or “the decision.” Instead, refer explicitly to people, organizations, actions, and the date mentioned.

Here’s an example of the format you should follow:

Example input:
Headline: "Amazon rainforest protections weakened in new Brazilian legislation"  
People mentioned: Luiz Inácio Lula da Silva; Marina Silva  
Organizations mentioned: Brazilian Congress; Greenpeace  
Date: June 4, 2023

Example output:
  1. What changes were made to Amazon rainforest protections in the Brazilian legislation passed on June 4, 2023?
  2. Who in the Brazilian Congress backed the bill weakening Amazon protections on June 4, 2023?
  3. Why did Luiz Inácio Lula da Silva support the rainforest law approved on June 4, 2023?
  4. How did Greenpeace respond to the Brazilian legislation passed on June 4, 2023?
  5. When did Marina Silva begin opposing the law on Amazon deforestation passed on June 4, 2023?
  6. Where in the Amazon could deforestation increase after the law passed on June 4, 2023?
  7. Is the Brazilian Congress facing backlash for passing weaker Amazon protections on June 4, 2023?
  8. Could Luiz Inácio Lula da Silva have blocked the law passed on June 4, 2023 that weakens rainforest protections?

Now, generate your 8 search-style questions using the same tone and level of clarity.

Here is the information you will use to generate the questions:

Headline: "{row.get('Headline', '')}"

People mentioned: {row.get('V2Persons', '')}

Organizations mentioned: {row.get('V2Organizations', '')}

Date: {row.get('date', '')}
\end{promptbox}

\subsection{Query translation}
\label{subsec:query-translation}

We created a simple translation query to translate queries either to Spanish, French or Portuguese using \texttt{gpt-4o-mini} in temperature 0.

\begin{promptbox}[System message]
Translate the following question into {target_lang}. Only return the translated question.
\end{promptbox}

\begin{promptbox}[User message]
{question} 
\end{promptbox}

\subsection{Domain categorisation}
\label{subsec:domain-category}

We created a query to categorise domains across the full taxonomy of Cloudflare categories; we used a temperature of 0.3 and \texttt{gpt-4o-mini} to provide the classifications.

\begin{promptbox}[System message]
You are a helpful and precise classification assistant.

I will provide you with the **domain name** and the **landing page text** of a website. Your task is to classify the content of the page into **exactly one** of the predefined categories listed below.

Please follow these rules:
- Base your decision **only on the content of the page text**, not the domain name.
- Choose the **single most appropriate** category, even if the content could potentially belong to more than one.
- If the page does not clearly fit into any category, choose **"Other"**.
- Respond with only the category name, **without any explanation or additional text**.

Here are the categories:

1. **Social Media**: Platforms for social networking, user-generated content, or community interaction.
2. **News & Media**: News outlets, media organizations, or content focused on current events.
3. **Business & Economy**: Domains related to business, finance, economics, science, or technology in a professional or institutional context, and sales.
4. **Education**: Educational content not covered by Science, Technology, or formal academic institutions.
5. **Entertainment**: General entertainment content.
6. **Government & Politics**: Government services, policies, political institutions, or political commentary.
7. **Personal Blogs**: Personal blogs, diaries, and vlogs.
8. **Other**: Use only if the page does not fit into any of the above categories.

Respond with only one category name from the list above.
\end{promptbox}

\begin{promptbox}[User prompt]
Please classify the following content:

Domain: `{row['url_domain']}`
Text:
{text[:3000]}
\end{promptbox}

%% file: appendix/extra/systems_and_settings.tex
\begin{table}[H]
  \centering
  \small
  \caption{\textbf{Systems and settings used for LLM-assisted search and conventional search.} For each system, we report when relevant the exact model string, knowledge cutoff, web search tool, and how we set the location of each query. $^\dagger$\, indicates information from an external source we could not directly confirm from documentation.}
  \label{tab:appx_systems_settings}
  \begin{tabularx}{\linewidth}{
      >{\raggedright\arraybackslash}p{3.3cm}
      >{\raggedright\arraybackslash}p{1.8cm}
      >{\raggedright\arraybackslash}X
      >{\raggedright\arraybackslash}X}
    \toprule
    System and model string & Knowledge cutoff & Web search configuration & Location targeting \\
    \midrule
    Grok 3 \newline (\texttt{grok-3-latest})
      & Nov.\ 2024--Feb.\ 2025$^\dagger$
      & Live Search tool available, model determines when to call it
      & Country passed as ISO code in web source parameters \\
    \addlinespace[5pt]
    GPT-4o mini \newline (\texttt{gpt-4o-mini-\allowbreak search-preview-\allowbreak 2025-03-11})
      & Oct.\ 2023
      & Web search tool called on every prompt, with search context set to medium
      & Country passed as ISO code in the approximate location parameters \\
    \addlinespace[5pt]
    Claude 3.5 Haiku \newline (\texttt{claude-3-5-\allowbreak haiku-20241022})
      & July 2024
      & Web search tool \texttt{web\_search\_20250305}, model determines when to call it
      & Country passed as ISO code in user location parameter \\
    \addlinespace[5pt]
    Perplexity Sonar \newline (\texttt{sonar})
      & Not published
      & Web search on every prompt, with low context size
      & Country passed as ISO code in the search location parameters \\
    \addlinespace[5pt]
    Google, Bing, and Yahoo! \newline via DataForSEO SERP API
      & n/a
      & Query submitted as the search keyword, top $10$ organic results retrieved, device
      & Country specified using the DataForSEO location code \\
    \bottomrule
  \end{tabularx}
\end{table}

%% file: appendix/extra/query_examples.tex
{\small
\begin{xltabular}{\linewidth}{l l >{\raggedright\arraybackslash}p{3.6cm} >{\raggedright\arraybackslash}p{3cm} >{\raggedright\arraybackslash}p{1.5cm} >{\raggedright\arraybackslash}X}
  \caption{\textbf{Example queries and GDELT metadata.} Each row represents one event, its country, its original headline, and any GDELT entity associated with it. The resulting query uses the headline, associated entities, and date to ensure there is no overlap between events.}
  \label{tab:appx_query_examples} \\
  \toprule
  Location & Type & Headline (lower-case) & Entities & Date & Selected query \\
  \midrule
  \endfirsthead

  \multicolumn{6}{l}{\textit{Table \thetable\ continued}} \\
  \toprule
  Location & Type & Headline (lower-case) & Entities & Date & Selected query \\
  \midrule
  \endhead

  \midrule
  \multicolumn{6}{r}{\textit{Continued on next page}} \\
  \endfoot

  \bottomrule
  \endlastfoot

  BR & Could & two arrested in brazil for plotting attack at lady gagas rio concert police report & Lady Gaga & 04 May 2025 & Could this incident affect Lady Gaga's future concerts in Brazil after the May 4, 2025 threat? \\
  CO & How & un warns of ongoing tragedy as indigenous groups in colombia face extinction & Scott Campbell; Gustavo Petro; Associated Press; Armed Forces Of Colombia; United Nations & 21 May 2025 & How has the Colombian government, led by Gustavo Petro, responded to the UN's concerns about indigenous groups on May 21, 2025? \\
  VE & Is & judge bars deportations of venezuelans from south texas under 18thcentury wartime law & Nicholas Riccardi; Donald Trump; Fernando Rodriguez Jr; Mark Sherman; Associated Press & 01 May 2025 & Is the Associated Press covering the implications of the judge's ruling on Venezuelans as of May 1, 2025? \\
  VE & What & trump administration asks supreme court to strip legal protections from 350000 venezuelan migrants & Donald Trump; Barack Obama; Edward Chen; Justice Department; Supreme Court & 02 May 2025 & What legal protections are being targeted for Venezuelan migrants by the Trump administration on May 2, 2025? \\
  BR & When & braço armado do hamas anuncia que libertará refém americanoisraelense nesta segunda & Her Office & 12 May 2025 & When is the expected release of the American-Israeli hostage scheduled to take place on May 12, 2025? \\
  MX & Where & los ángeles azules claudia sheinbaum ordena a guardia nacional atender robo a camión en carretera méxicopuebla & National Guard & 13 May 2025 & Where exactly did the truck robbery on May 13, 2025, take place on the Mexico-Puebla highway? \\
  CI & Who & china hosts latin american caribbean nations amid us trade war talks & Michael Perry; Ella Cao; Laurie Chen; European Union; Reuters & 12 May 2025 & Who are the key Chinese officials meeting with Latin American leaders amid US trade war talks on May 12, 2025? \\
  MX & Why & things to do in the quadcities cinco de mayo kentucky derby and other weekend events & Peter Pan; Victor Granados; Kentaco Derby & 01 May 2025 & Why is the Kentaco Derby a highlight of the weekend events in the Quad Cities on May 1, 2025? \\
\end{xltabular}
}

%% file: appendix/extra/query_design.tex
\begin{table}[H]
  \centering
  \small
  \caption{\textbf{Summary statistics for the filtering pipeline,} by country. For each country, we began with an initial set of 400 candidate events from GDELT. We then retained the number of events shown below after filtering. Sources indicate how many URLs were collected from language models and search engines altogether. News \& Media sources indicate how many sources were classified as News \& Media using the Cloudflare taxonomy.}
  \label{tab:appx_query_language_by_country}
  \begin{tabular}{llrrr}
    \toprule
    Country & Local language & Retained events & Sources & News \& Media sources \\
    \midrule
    AR & Spanish & 211 & 44,913 & 23,499 \\
    BR & Portuguese & 305 & 60,331 & 27,721 \\
    CL & Spanish & 152 & 36,720 & 17,761 \\
    CO & Spanish & 299 & 60,001 & 30,429 \\
    HT & French & 265 & 55,023 & 29,117 \\
    MX & Spanish & 232 & 50,701 & 28,241 \\
    PE & Spanish & 108 & 21,305 & 11,207 \\
    SV & Spanish & 140 & 34,173 & 22,676 \\
    VE & Spanish & 189 & 39,276 & 24,554 \\
    \bottomrule
  \end{tabular}
\end{table}

%% file: appendix/extra/worked_example.tex
\begin{table}[H]
  \centering
  \small
  \caption{\textbf{Example response for one random event in the dataset.} We show the headline and query for event \textit{\#1245080134}, about a detained veteran in Venezuela, along with Grok 3's response and its first six cited sources. Sources are included whether or not they are cited directly in response text.}
  \label{tab:appx_worked_example}
  \begin{tabularx}{\linewidth}{lX}
    \toprule
    Event ID & 1245080134 \\
    Event country & VE \\
    Headline & venezuela frees us air force veteran considered wrongfully detained his family says \\
    Selected query & Who advocated for the release of the wrongfully detained veteran in Venezuela, according to his family? \\
    \midrule
    Grok 3 reply (excerpt) & \begin{minipage}[t]{0.8\textwidth}According to the family of Joseph St. Clair, a U.S. Air Force veteran wrongfully detained in Venezuela, they expressed gratitude to President Donald Trump and his administration for helping to secure his release. Additionally, posts found on X mention the involvement of Special Envoy Richard Grenell in negotiations and secret talks with Venezuelan officials that contributed to St. Clair's release on May 20, 2025. The family's statements are supported by multiple news sources, including Fox News and CNN, as ...\end{minipage} \\
    \midrule
    Cited sources (first six) & \begin{minipage}[t]{0.8\textwidth}\begin{itemize}[leftmargin=*,nosep]
    \item \url{https://www.foxnews.com/politics/trump-administration-secures-release-us-veteran-held-venezuela-family-says}
    \item \url{https://www.nbcnews.com/news/latino/venezuelan-rare-major-surgery-was-deported-el-salvador-prison-family-n-rcna215416}
    \item \url{https://www.nytimes.com/2025/05/20/world/americas/venezuela-us-air-force-veteran-release.html}
    \item \url{https://apnews.com/article/venezuela-us-national-freed-detention-272f5262f22808ae11ac0e06183a3219}
    \item \url{https://www.washingtonpost.com/world/2025/05/20/venezuela-maduro-air-force-veteran-released/}
    \item \url{https://m.economictimes.com/news/international/world-news/wrongfully-detained-in-venezuela-us-air-force-veteran-joe-st-clair-returns-home-to-family/amp_podcast/121315347.cms}
    \end{itemize}\end{minipage} \\
    \bottomrule
  \end{tabularx}
\end{table}

%% file: appendix/tables/outlet_collision_by_user.tex
\begin{table}[H]
  \centering
  \caption{\textbf{Effective number of News \& Media outlets for each language model and search engine,} by user type. Higher values indicate a more diverse set of outlets. Results are reported for within-country and across-country comparisons, with 95\% confidence intervals.}
  \label{tab:appx_outlet_collision_by_user}
  \setlength{\tabcolsep}{5pt}
  \begin{tabular}{
    @{}
    >{\raggedright\arraybackslash}p{1.5cm}
    l
    S[table-format=3.1]
    c
    S[table-format=3.1]
    c
    @{}
  }
    \toprule
    & & \multicolumn{2}{c}{Within country} & \multicolumn{2}{c}{Across countries} \\
    \cmidrule(lr){3-4} \cmidrule(lr){5-6}
    User & System & {Eff.\ outlets ($N_{su}$)} & 95\% CI & {Eff.\ outlets ($N_{su}$)} & 95\% CI \\
    \midrule
    \multirow{7}{*}{Local} & Claude 3.5 Haiku & 29.2 & \CI{26.3}{32.6} & 127.6 & \CI{110.1}{147.9} \\
     & GPT-4o & 16.1 & \CI{14.9}{17.4} & 33.9 & \CI{30.5}{37.9} \\
     & Grok 3 & 42.3 & \CI{39.3}{45.8} & 147.5 & \CI{137.5}{158.7} \\
     & Perplexity & 18.4 & \CI{16.7}{20.2} & 46.9 & \CI{42.4}{51.8} \\
    \cmidrule(lr){2-6}
     & Google & 24.3 & \CI{22.6}{26.2} & 76.5 & \CI{69.5}{84.6} \\
     & Bing & 53.6 & \CI{49.8}{57.7} & 156.2 & \CI{142.8}{171.3} \\
     & Yahoo! & 44.9 & \CI{41.6}{48.4} & 142.5 & \CI{131.4}{154.8} \\
    \midrule
    \multirow{7}{*}{Diaspora} & Claude 3.5 Haiku & 31.1 & \CI{28.6}{33.8} & 91.2 & \CI{82.9}{100.4} \\
     & GPT-4o & 15.7 & \CI{14.5}{17.1} & 29.5 & \CI{26.5}{32.9} \\
     & Grok 3 & 39.8 & \CI{36.9}{42.9} & 148.8 & \CI{137.9}{161.0} \\
     & Perplexity & 17.4 & \CI{15.8}{19.0} & 44.2 & \CI{39.9}{48.8} \\
    \cmidrule(lr){2-6}
     & Google & 31.3 & \CI{29.0}{33.9} & 79.8 & \CI{73.6}{86.8} \\
     & Bing & 49.4 & \CI{46.4}{52.6} & 128.3 & \CI{118.6}{138.8} \\
     & Yahoo! & 40.3 & \CI{37.8}{42.9} & 115.1 & \CI{106.6}{124.9} \\
    \midrule
    \multirow{7}{*}{Tourist} & Claude 3.5 Haiku & 40.1 & \CI{36.4}{44.5} & 68.6 & \CI{60.8}{78.3} \\
     & GPT-4o & 13.0 & \CI{11.9}{14.3} & 16.7 & \CI{15.1}{18.8} \\
     & Grok 3 & 47.7 & \CI{45.2}{50.4} & 71.0 & \CI{66.6}{75.8} \\
     & Perplexity & 38.7 & \CI{34.8}{43.5} & 118.1 & \CI{105.0}{134.8} \\
    \cmidrule(lr){2-6}
     & Google & 65.2 & \CI{59.8}{71.2} & 127.9 & \CI{117.9}{139.2} \\
     & Bing & 78.4 & \CI{72.6}{84.9} & 142.8 & \CI{130.6}{156.4} \\
     & Yahoo! & 78.6 & \CI{72.6}{85.2} & 138.2 & \CI{124.4}{154.3} \\
    \midrule
    \multirow{7}{*}{Foreigner} & Claude 3.5 Haiku & 32.7 & \CI{30.6}{34.8} & 48.0 & \CI{44.5}{52.0} \\
     & GPT-4o & 11.8 & \CI{10.6}{13.1} & 13.7 & \CI{12.3}{15.4} \\
     & Grok 3 & 44.6 & \CI{42.3}{47.2} & 67.1 & \CI{63.0}{71.9} \\
     & Perplexity & 34.7 & \CI{30.7}{39.1} & 118.1 & \CI{104.8}{135.1} \\
    \cmidrule(lr){2-6}
     & Google & 79.9 & \CI{73.2}{87.1} & 124.0 & \CI{113.2}{136.3} \\
     & Bing & 87.1 & \CI{81.0}{93.7} & 132.5 & \CI{121.0}{145.6} \\
     & Yahoo! & 82.6 & \CI{77.3}{88.7} & 127.5 & \CI{116.9}{139.6} \\
    \bottomrule
  \end{tabular}
\end{table}

%% file: appendix/tables/simple_news_vs_average_models.tex
\begin{table}[H]
  \centering
  \caption{\textbf{Difference in share of domestic news relative to the average search engine,} by user type. Domestic news are News \& Media domains from the same country as the event. We compute the share of domestic news per response, and report the mean difference in percentage points (p.p.) in average across all responses to avoid over-representing responses with more sources. Positive values indicate a higher share of domestic news. Values with $^*$ indicate a confidence interval that excludes zero.}
  \label{tab:appx_simple_news_vs_average_models}
  \setlength{\tabcolsep}{7pt}
  \renewcommand{\arraystretch}{1.08}
  \begin{tabular}{@{}llS[table-format=+2.1]l@{}}
    \toprule
    User & System & \multicolumn{1}{c}{Difference (pp)} & \multicolumn{1}{c}{95\% CI} \\
    \midrule

    Local & Claude 3.5 Haiku & -9.2$^*$ & \CI{-11.3}{-7.0} \\
    & GPT-4o & -3.6$^*$ & \CI{-5.3}{-1.8} \\
    & Grok 3 & -9.5$^*$ & \CI{-11.0}{-8.2} \\
    & Perplexity & -11.6$^*$ & \CI{-12.9}{-10.3} \\
    \midrule
    
    Diaspora & Claude 3.5 Haiku & -4.1$^*$ & \CI{-5.6}{-2.6} \\
    & GPT-4o & +1.5 & \CI{-0.1}{+3.1} \\
    & Grok 3 & +2.1$^*$ & \CI{+0.8}{+3.4} \\
    & Perplexity & -1.5$^*$ & \CI{-2.8}{-0.2} \\
    \midrule
    
    Tourist & Claude 3.5 Haiku & +1.0 & \CI{-0.5}{+2.5} \\
    & GPT-4o & +5.9$^*$ & \CI{+4.6}{+7.3} \\
    & Grok 3 & -7.6$^*$ & \CI{-8.6}{-6.7} \\
    & Perplexity & -3.7$^*$ & \CI{-4.8}{-2.5} \\
    \midrule
    Foreigner & Claude 3.5 Haiku & -0.4 & \CI{-1.2}{+0.5} \\
    & GPT-4o & +3.4$^*$ & \CI{+2.2}{+4.7} \\
    & Grok 3 & -2.3$^*$ & \CI{-3.1}{-1.5} \\
    & Perplexity & +1.8$^*$ & \CI{+0.7}{+2.9} \\
    \bottomrule
  \end{tabular}
\end{table}